\documentclass[acmtog]{acmart}

\setcopyright{none}
\renewcommand\footnotetextcopyrightpermission[1]{}
\usepackage{balance}
\usepackage{booktabs} 
\usepackage{listings}
\usepackage{color}
\usepackage[skip=2pt]{caption}
\usepackage{subcaption}
\usepackage{cleveref}
\usepackage{textcomp}

\def\BibTeX{{\rm B\kern-.05em{\sc i\kern-.025em b}\kern-.08emT\kern-.1667em\lower.7ex\hbox{E}\kern-.125emX}}

\DeclareCaptionFormat{captionformat}{\fontsize{7}{8}\selectfont#1#2#3}
\definecolor{dkgreen}{rgb}{0,0.6,0}
\definecolor{gray}{rgb}{0.5,0.5,0.5}
\definecolor{mauve}{rgb}{0.58,0,0.82}

\mathchardef\mhyphen="2D

\begin{document}
\title{Improving Deep Learning For Airbnb Search}
\author{Malay Haldar,  Prashant Ramanathan, Tyler Sax, Mustafa Abdool, Lanbo Zhang}
\author{Aamir Mansawala, Shulin Yang, Bradley Turnbull, Junshuo Liao}
\affiliation{
  \institution{Airbnb Inc.}
}
\email{malay.haldar@airbnb.com}

\renewcommand{\shortauthors}{Malay Haldar et al.}

\begin{abstract}
The application of deep learning to search ranking was one of the most impactful product improvements at Airbnb. But what comes next after you launch a deep learning model? In this paper we describe the journey beyond, discussing what we refer to as the ABCs of improving search: $\mathcal{A}$ for architecture, $\mathcal{B}$ for bias and $\mathcal{C}$ for cold start. For architecture, we describe a new ranking neural network, focusing on the process that evolved our existing DNN beyond a fully connected two layer network. On handling positional bias in ranking, we describe a novel approach that led to one of the most significant improvements in tackling inventory that the DNN historically found challenging. To solve cold start, we describe our perspective on the problem and changes we made to improve the treatment of new listings on the platform. We hope ranking teams transitioning to deep learning will find this a practical case study of how to iterate on DNNs.
\end{abstract}

%
%



\maketitle
\thispagestyle{empty}

\section{Introduction}
Airbnb is a two sided marketplace, bringing together hosts who own places to rent, with prospective guests from across the globe. The search ranking problem at Airbnb is to rank the places to stay, referred to as listings, in response to a query from the guest which typically consists of a location, number of guests and checkin/checkout dates. Transitioning to deep learning was a major milestone in the evolution of search ranking at Airbnb. Our account of the journey in ~\cite{kdd19} brought us in conversation with many industry practitioners, allowing us to exchange insights and critiques. One question that frequently followed such conversations: what next? We try to answer that in this paper.

The launch of deep learning for ranking was cause of much celebration, not only because of the gains in bookings it generated, but because of the change it brought to our roadmap ahead. The initial perception was that having Airbnb ranking on deep learning gave us access to this vast treasure trove of machine learning ideas, which only seemed to be growing each day. We could simply pick the best ideas from literature surveys, launch them one after another, and live happily ever after. But this proved to be the peak of optimism. The familiar pattern of descent into the valley of despair followed soon, where techniques with impressive success elsewhere proved quite neutral on our own application.

This lead to a complete revision of our strategy on how to iterate on deep learning beyond the first launch. In this paper we capture the major enhancements that followed the launch of the DNN described in ~\cite{kdd19}. In addition to delving into the core machine learning techniques themselves, we focus on the process and the reasoning that lead to the breakthroughs. With the bigger picture in view now, we value the lessons learnt on how to iterate on DNNs more than any individual technique. We hope those focused on applying deep learning in industry settings will find our experiences valuable. We open the discussion by taking a look at our efforts to improve the DNN architecture.

\section{Optimizing The Architecture} \label{conehead}
What is deep learning all about? Well, adding more layers. At least that was our na\"ive interpretation after reviewing the series of advances that ushered in the current deep learning era. But as we sought to replicate the benefits of scaling data and adding layers as summarized in ~\cite{Sun-2017-ICCV}, we met nothing but neutral test results. Trying to decipher why increasing layers was not showing any gains led us to borrow more ideas from the literature, like applying residual learning ~\cite{He-2016-CVPR} and batch normalization ~\cite{pmlr-v37-ioffe15}. Still, NDCG refused to budge in offline tests. Our takeaway from the exercise was that increasing layers was an effective technique for convolutional neural networks, but not necessarily for all DNNs. For fully connected networks like ours, two hidden layers were sufficient and model capacity was not our problem.

If deeper nets were not the right architecture for us, we hypothesized, more specialized architectures might be. So we tried architectures that could tackle interaction between query and listings more explicitly, like deep and wide ~\cite{deep_and_wide}, where query-listing feature crosses were added to the wide part. This was followed by variants of attention based networks from ~\cite{attention}. The intention there was to make the hidden layer derived from query features focus its attention on certain parts of the hidden layer derived from listing features. The short summary of those efforts is that they too failed to move the needle. 

In trying to import successful deep learning architectures to product application, what often gets lost in translation is that the success of an architecture is intricately tied to its application context. The reported performance gains of an architecture comes from addressing certain shortcomings of the baseline it is compared to. With the general lack of explainability of deep learning, it becomes difficult to infer exactly what shortcoming the new architecture is addressing and how. Determining whether those exact shortcomings are also plaguing the product at home, therefore, becomes a guesswork.

To improve our chances of success, we abandoned the \{download paper $\mapsto$ implement $\mapsto$ A/B test\} loop. Instead we decided to drive the process based on a very simple principle: users lead, model follows.

\subsection{Users lead, model follows} \label{userdriven}
\begin{figure}
\includegraphics[height=1.8in, width=2.5in]{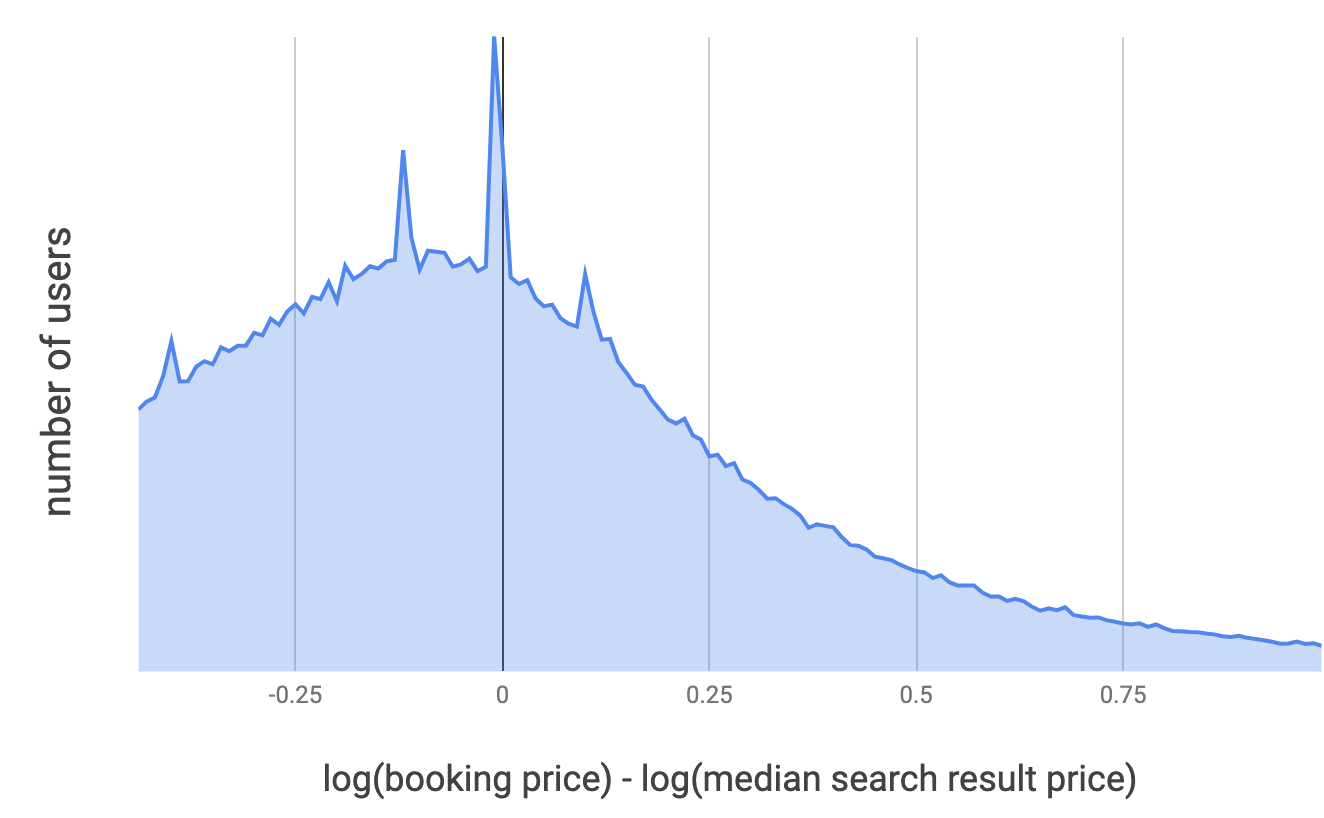}
\caption{X-axis shows how the price of the booked listing offsets from the median price of search results for a guest. Y-axis is the number of users corresponding to a price offset.}
\label{fig:impbkdiff}
\end{figure}
The idea here is to first quantify a user problem. Model tweaks come afterwards, and in response to the user problem.

Along those lines, we started with the observation that the series of successful ranking model launches described in ~\cite{kdd19} were not only associated with an increase in bookings, but also a reduction in the average listing price of search results. This indicated the model iterations were moving closer to the price preference of guests, which was lower than what the previous models had estimated. We suspected that even after the successive price reductions, there was likely a gap between the model\textquotesingle s choice of prices and what guests preferred. To quantify this gap we looked at the distribution of the difference between the median price of search results seen by a guest and the price of the listing that the guest booked. The difference is computed after taking $log$ of the prices, as price follows a log-normal distribution. Figure~\ref{fig:impbkdiff} plots how the difference is distributed.

Our expectation was that the booked price would be symmetrically distributed around the median price of search results, and resemble a normal distribution centered at zero. Instead it was heavy on the negative side, indicating a skewed guest preference towards lower prices. This gave us a concrete user problem to investigate: whether lower priced listings that were closer to guests\textquotesingle ~preferred prices needed to be ranked higher.

Given two ordinary listings with everything else equivalent, our intuitive understanding was that guests would prefer the more economical listing. Did our ranking model truly understand this {\it cheaper is better} principle? We were not completely sure.

\subsection{Enforcing {\it Cheaper Is Better}} \label{pricepenalty}
The reason we lacked clarity on how the model was interpreting the listing price was because it was a DNN. Familiar tools like inspecting the corresponding weights in logistic regression models or plotting partial dependence graphs for GBDT models were not as effective in the DNN context any longer.

To make price more interpretable we applied the following changes:
\begin{itemize}
\item Removed price as an input feature to the DNN. We represent this modified DNN as $DNN_{\theta}(u, q, l_{no\_price})$. Here $\theta$ are the DNN parameters, $u$ user features, $q$ query features and $l_{no\_price}$ listing features with the exception of price.
\item Represent the final output of the model as
\begin{equation}
\begin{aligned}[t]
DNN_{\theta}(u, q, l_{no\_price}) - tanh(w*\mathcal{P} + b)
\end{aligned}
\label{tanheq}
\end{equation}
with $w$ and $b$ as additional parameters learnt using back propagation, and
\begin{equation*}
\begin{aligned}[t]
\mathcal{P} = log(\frac{1 + price}{1 + price_{median}})
\end{aligned}
\end{equation*}
Here $price$ is the raw price feature and $price_{median}$ is a constant computed from the median of the logged listing prices.
\end{itemize}

\begin{figure}
\includegraphics[height=1.4in, width=2.1in]{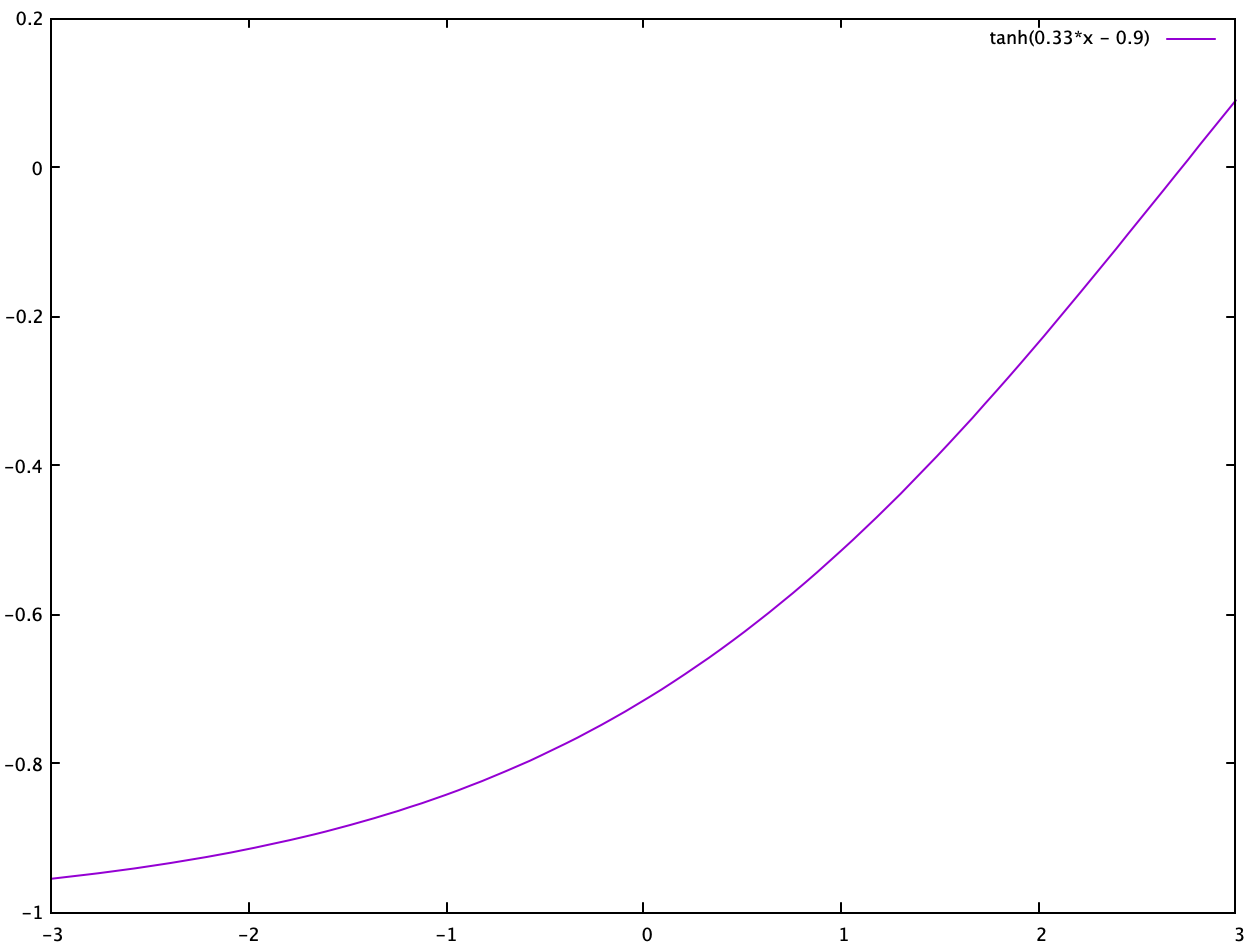}
\caption{X-axis is normalized price feature. Y-axis is the value of the $tanh$ term in Equation~\ref{tanheq}.}
\label{fig:tanh}
\end{figure}

The $-tanh()$ term allowed us to enforce {\it cheaper is better} by monotonically decreasing the output score with respect to increasing price. The readily interpretable $w$ and $b$ parameters allowed us to plot out the precise effect of price. For the learned values of the parameters, $w = 0.33$ and $b = -0.9$, the plot is shown in Figure~\ref{fig:tanh} over the typical range of $\mathcal{P}$ encountered during ranking.

When tested online as an A/B experiment against the two hidden layer DNN from ~\cite{kdd19}, average price of search results dropped by $-5.7\%$, in confirmation with offline analysis. But the interpretability of price came at a heavy cost as bookings dropped by $-1.5\%$. Our hypothesis was that price interacted heavily with other features. Isolating price away from the model resulted in under fitting. This hypothesis was supported by the fact that both training and test NDCG had declined.

\subsection{Generalized Monotonicity} \label{hmon}
\begin{figure}
\includegraphics[height=1.5in, width=2.2in]{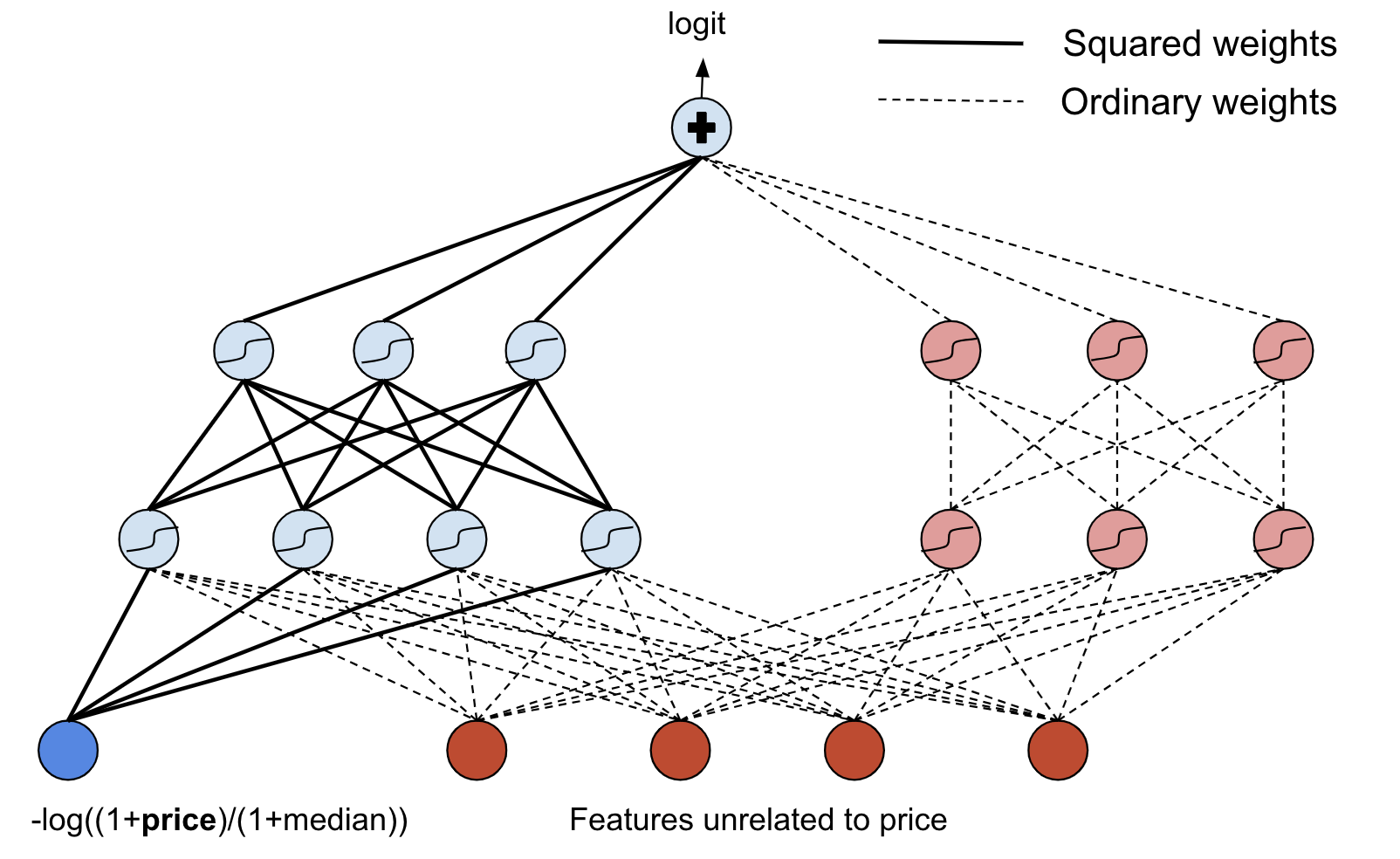}
\caption{DNN architecture partially monotonic w.r.t price. Bold solid lines indicate weights that are squared, dashed lines indicate ordinary weights.}
\label{fig:monoarch}
\end{figure}

To retain the {\it cheaper is better} intuition in the model, but allow price to interact with rest of the features, we started investigating DNN architectures that were monotonic with respect to some of its inputs. Lattice networks described in ~\cite{lattice} presented an elegant solution to the problem. But pivoting our entire system to lattice networks presented a big challenge and we sought a mechanism that was less disruptive. So we constructed the architecture shown in Figure~\ref{fig:monoarch} that doesn't depend on any specialized computational nodes other than those natively present in Tensorflow\textsuperscript{TM}. We discuss the step by step construction of the architecture, ensuring all paths from the input price node to the final output are monotonic with respect to price:
\begin{itemize}
\item We feed $-\mathcal{P}$ as input to the DNN which is monotonically decreasing w.r.t price. 
\item At the input layer, instead of multiplying $-\mathcal{P}$ by the weight, we multiply by the square of the weight. Since $-w^2*\mathcal{P} + b$ is monotonically decreasing for any real values of $w$ and $b$, the inputs to the first hidden layer are always monotonically decreasing w.r.t price. 
\item For the hidden layers we use the tanh activation which preserves the monotonic property.
\item Given $f_0(x)$ and $f_1(x)$, two monotonically decreasing functions of $x$, ${w_{0}}^2*f_0(x) + {w_{1}}^2*f_1(x) + b$ is also monotonically decreasing w.r.t $x$ where $w_0$ and $w_1$ can be arbitrary real weights. We use this property in the second hidden layer and the output layer where all the weights are squared. These are represented as the bold solid lines for the second hidden layer and output layer in Figure~\ref{fig:monoarch}.
\item A subnet which neither has price as input nor any of the monotonicity constraints is added to allow unconstrained interaction between the rest of the features.
\end{itemize}
 
In spite of being more flexible than the architecture described in section ~\ref{pricepenalty}, when tested online the results were very similar, resulting in a booking drop of $-1.6\%$. Like its predecessor, the architecture enforced the model output was monotonically decreasing w.r.t price under all circumstances. The failure of this architecture suggested that monotonicity with regard to price was too strict a constraint.

\subsection{Soft Monotonicity}
While the architecture described in section ~\ref{hmon} revealed how versatile DNNs could be in supporting model constraints, it also taught us another trait of DNNs: that they behaved just like another star engineer on the team. Given a problem and left to their own devices, they usually came up with a reasonable solution. But force them to go some direction, and disaster will quickly follow. So in our next iteration, we decided to manage the DNN by setting context, not control. Instead of enforcing that the model output be monotonic with respect to price, we added a soft hint that cheaper was better.

Ordinarily, each training example consisted of a pair of listings, one booked and the other not booked. Applying the DNN to the features of the two listings generated the corresponding logits, and the loss was defined as shown in Table~\ref{ordinaryloss}.
\begin{table}
\begin{lstlisting}[basicstyle=\tiny]
def get_loss_op(positive_logits, negative_logits):
  """ Create the loss op to be minimized.
  """
  logit_diffs = positive_logits - negative_logits
  xentropy = tf.nn.sigmoid_cross_entropy_with_logits(
      labels=tf.ones_like(logit_diffs),
      logits=logit_diffs)
  loss = tf.reduce_mean(xentropy)
  return loss

# Booked listings as positives, not booked as negatives
loss = get_loss_op(booked_logits, not_booked_logits)
\end{lstlisting}
\caption{TensorFlow\textsuperscript{TM} code for pairwise booking loss.}
\label{ordinaryloss}
\end{table}

To add the price hint, we introduce a second label for each training example, indicating which listing in the pair has a lower price and which one the higher. The loss is then modified as shown in Table~\ref{priceloss}. The $alpha$ hyperparameter gives a way to control whether we want the results to be sorted by relevance or price.
\begin{table}
\begin{lstlisting}[basicstyle=\tiny]
# Booked listings as positives, not booked as negatives
booking_loss = get_loss_op(booked_logits, not_booked_logits)
# Lower priced listings as positives, higher priced listing as negatives
price_loss = get_loss_op(lower_price_logits, higher_price_logits)
# Total loss a linear combination with a hyperparameter
loss = alpha*booking_loss + (1 - alpha)*price_loss
\end{lstlisting}
\caption{TensorFlow\textsuperscript{TM} code with price loss added.}
\label{priceloss}
\end{table}

To test the idea, we adjusted the $alpha$ hyperparameter to the minimum value such that in offline tests we got the same NDCG as the baseline model. This allowed us to push the {\it cheaper is better} intuition as far as possible without hurting relevance, at least when measured offline. In the online A/B test, we observed a reduction of $-3.3\%$ in average price of search results. But also a drop of $-0.67\%$ in bookings. The offline analysis suffered from the limitation that it only evaluated re-ranking the top results available in logs. During the online test, applying the newly trained model to the entire inventory revealed the true cost of adding the price loss as part of the training objective. 
 
\subsection{Putting Some ICE}
The trail of disaster from the price lowering experiments left us in a paradoxical state: the listing prices in search results seemed higher than what guests preferred, but pushing prices down made guests unhappy. To understand where the new models were falling short, it was necessary to compare how the baseline model was utilizing the price feature, but that was shrouded in the lack of interpretability of the fully connected DNN. As mentioned previously, concepts like partial dependence plots were not useful since they relied on the assumption that a given feature\textquotesingle s influence on the model was independent of other features. This was simply not true in the case of DNNs. Attempts to plot partial dependence for price produced gently sloping straight lines, suggesting the DNN had some mild linear dependence on price, which was in contradiction with everything else we knew.

\begin{figure}
\includegraphics[height=1.5in, width=2.2in]{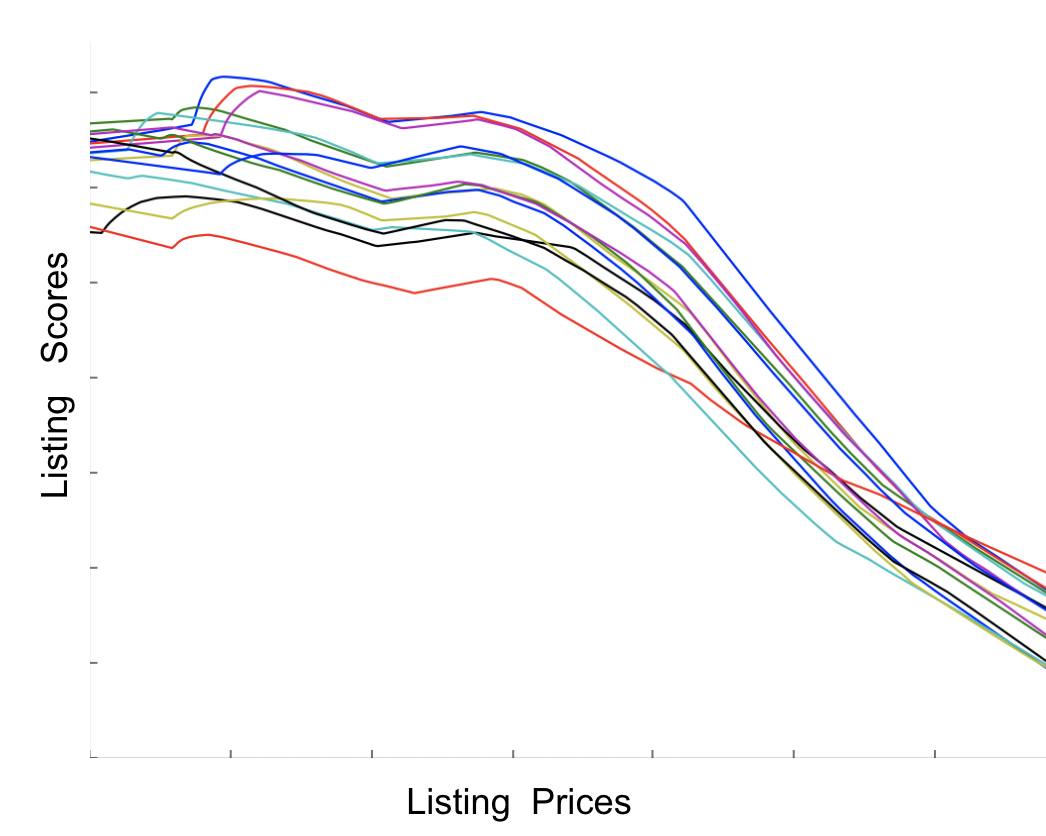}
\caption{ICE plot of listing scores (y-axis) vs listing prices (x-axis). Each curve represents a listing in the search result.}
\label{fig:icebonehead}
\end{figure}

To make progress, we scaled down the problem of DNN interpretability. Instead of trying to make general statements about how price influenced the DNN, we focused on interpreting one search result at a time. Borrowing the idea of individual conditional expectation (ICE) plots from ~\cite{iceplot}, we took listings from a single search result, swept across the price range while keeping all other features invariant, and constructed plots of the model score. An example plot is shown in Figure~\ref{fig:icebonehead}. The plots suggested that the fully connected two layer DNN from ~\cite{kdd19} already understood cheaper was better. Repeating the ICE analysis on a collection of randomly selected searches from the logs further strengthened this conclusion. By trying to force price down further, the failed architectures were compromising on quality. 

\subsection{Two Tower Architecture} \label{twotower}
Going back to Figure~\ref{fig:impbkdiff}, guests were clearly sending a message through that plot. But the architectures intent on trading relevance for price were interpreting the message incorrectly. A reinterpretation of Figure~\ref{fig:impbkdiff} was in order. And the reinterpretation had to align with price, as well as relevance.

\begin{figure}
\includegraphics[height=1.6in, width=2.5in]{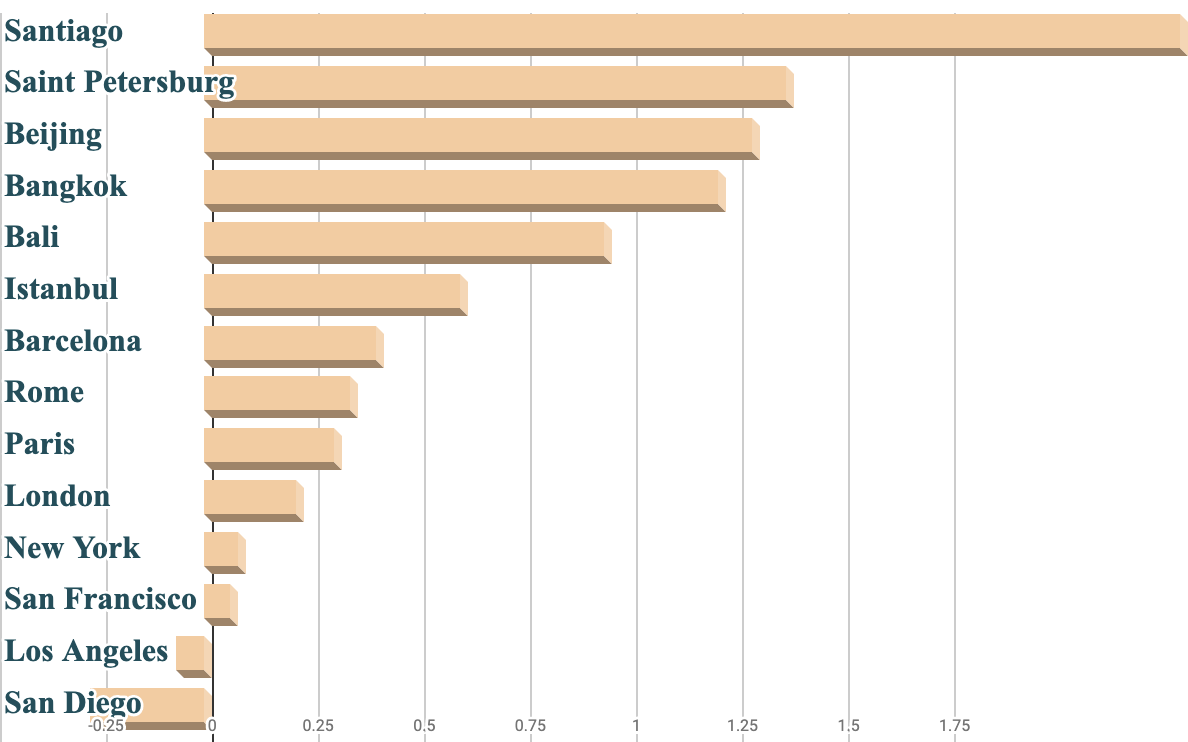}
\caption{Average difference between the median price of search results for a guest and the price of the booked listing, split by cities}
\label{fig:avgcitydiff}
\end{figure}

Such an alternate explanation for Figure~\ref{fig:impbkdiff} surfaced when we took the difference between the median price of search results for a guest and the price at which they booked, and computed averages grouped by cities. As expected, there was variance across the cities. But the differences were much larger for tail cities, compared to head cities. The tail cities were often located in developing markets as well. Figure~\ref{fig:avgcitydiff} shows the average difference between the median price of search results and the booked price for some selected cities.

This gave rise to the hypothesis that the DNN behind Figure~\ref{fig:impbkdiff} was suffering from the tyranny of the majority, focusing on price-quality tradeoffs that were tuned for the most popular locations that dominated bookings. Generalizing those tradeoffs to the tail queries was not working as well and the model was failing to adapt to local conditions. 

The hypothesis played well with another observation about the features feeding the DNN. Given the DNN was trained using a pairwise loss, features that differed across the two listings forming the pair seemed to have most of the influence. Query features, that were common across the pair, seemed to have little influence and dropping them impacted the NDCG minimally.

The new thinking was that the model had ample understanding of {\it cheaper is better}, what it was missing was the notion of {\it the right price for a trip}. Grasping this notion involved paying closer attention to query features like the location, instead of discriminating purely based off listing features.

This inspired the next revision of the architecture which consisted of two towers, similar to ~\cite{krichene2018efficient}. The first tower, fed by the query and user features, generated a $100 \mhyphen d$ vector which conceptually represented the ideal listing for the query-user combination. The second tower constructed a $100 \mhyphen d$ vector from the listing features. The euclidean distance between the two vectors was used as a measure of how far the given listing was from the ideal listing for the query-user.

\begin{figure}
\includegraphics[height=1.9in, width=3.2in]{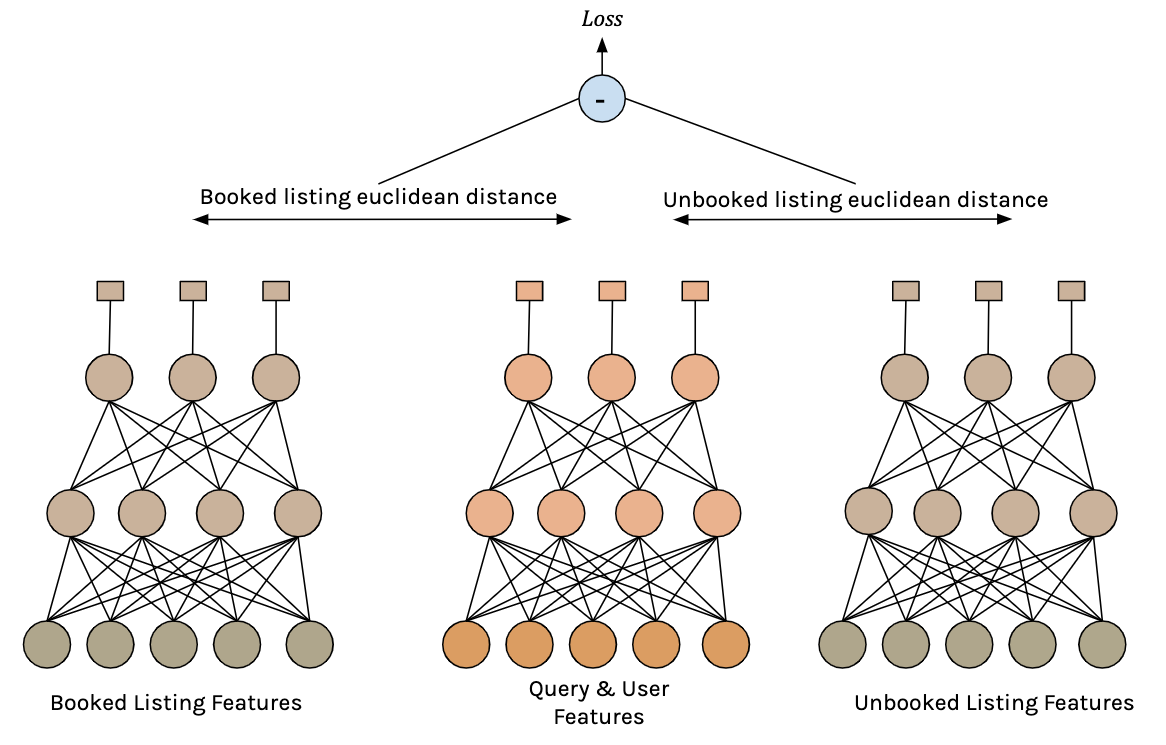}
\caption{Pairwise loss computation for training the two tower architecture.}
\label{fig:coneheadarch}
\end{figure}

Training examples consisted of pairs of listings: one booked, the other not booked. Loss was defined by how close the not booked listing was to the ideal, compared to the booked listing. Training of the two towers therefore brought the booked listing in the pair closer to the ideal, while pushing the not booked listing away. This is similar to the triplet loss introduced in ~\cite{triplet}. The main difference here is that instead of training on triples, we only have pairs of listings, and the missing anchor listing in the triple is learnt automatically by the query-user tower. The pairwise training of the query and listing towers is depicted in Figure~\ref{fig:coneheadarch}. Table~\ref{coneheadloss} shows the abstracted Tensorflow\textsuperscript{TM} code for the architecture. The actual implementation is slightly different to optimize for training speed.  

\begin{table}
\begin{lstlisting}[basicstyle=\tiny]
import tensorflow as tf

def get_tower(features, w0, b0, w1, b1):
  '''Two fully connected hidden layers producing a 100-d vector''' 
  h1 = tf.nn.tanh(tf.matmul(features, w0) + b0)
  h2 = tf.nn.tanh(tf.matmul(h1, w1) + b1)
  return h2

def get_distance_to_ideal(query_vec, listing_vec):
  '''Euclidean distance of the listing hidden layer to the query
     hidden layer. In practice minimizing sum of squared diff is
     equivalent to minimizing the Euclidean distance.'''
  sqdiff = tf.math.squared_difference(query_vec, listing_vec)
  logits = tf.math.reduce_sum(sqdiff, axis=1)
  return logits

def pairwise_loss(query_features,
                  booked_listing_features,
                  not_booked_listing_features):
  qvec = get_tower(query_features,
      query_w0, query_b0, query_w1, query_b1)
  booked_vec = get_tower(booked_listing_features,
      listing_w0, listing_b0, listing_w1, listing_b1)
  not_booked_vec = get_tower(listing_features,
      listing_w0, listing_b0, listing_w1, listing_b1)
                        
  booked_distance = get_distance_to_ideal(qvec, booked_vec)
  not_booked_distance = get_distance_to_ideal(
      qvec, not_booked_vec)
  distance_diff = not_booked_distance - booked_distance
  # Push the not booked away and the booked closer to
  # the ideal by increasing relative distance in between.
  xentropy = tf.nn.sigmoid_cross_entropy_with_logits(
    labels=tf.ones_like(logit_diffs),
    logits=logit_diffs)
  loss = tf.reduce_mean(xentropy)
  return loss


\end{lstlisting}
\caption{Abstracted TensorFlow\textsuperscript{TM} code for the two tower architecture.}
\label{coneheadloss}
\end{table}

\subsection{Test Results}
When tested online in an A/B experiment against the fully connected two layer DNN from ~\cite{kdd19}, the two tower architecture recorded a bookings gain of $+0.6\%$. The gain was driven by increased ease of search, as the NDCG computed online improved by $+0.7\%$. Although the two tower architecture was not directly aiming to lower prices, we observed a drop of $-2.3\%$ in average price of search results as a side effect of the increased relevance. The increase in bookings more than offset the effect of the price drop on revenue, resulting in an overall increase of $+0.75\%$.

In addition to improving the quality of the results, the two tower architecture allowed us to optimize the latency of scoring the DNN online as well. For the fully connected architecture, evaluating the first hidden layer contributed the largest component of scoring latency. The computation complexity of evaluating the first hidden layer can be expressed as $O(H*(Q + L))$ where $Q$ is the number of query and user features independent of the listing, $L$ the number of listing dependent features, and $H$ the number of hidden units for the first layer. For evaluating a search result set with N listings, the total complexity can be expressed as $O(N*H*(Q+L))$.

Of the two towers in the new architecture, the query tower was independent of listings. This allowed scoring that tower exactly once for the entire search result set, and only evaluating the listing dependent tower for each listing. The computational complexity of the first hidden layer reduced to $O(N*H_l*L + H_q*Q)$, where $H_l$ and $H_q$ are the number of hidden units in the listing and query towers. When tested online this resulted in a $-33\%$ reduction in the 99th percentile scoring latency.

\subsection{Architecture Retrospective}
Even as we celebrated the success, the doubt that invariably followed the launch of a DNN iteration crept up. Was the architecture working as intended, or did the DNN stumble onto something else unintended? The impenetrable nature of DNNs had made answering such doubts extremely hard in the past. But given that the intuition for the two tower architecture was developed in response to a user problem, we could use those intuitions now to get a better understanding of how the DNN was functioning. 

\begin{figure}
\includegraphics[height=1.5in, width=2.2in]{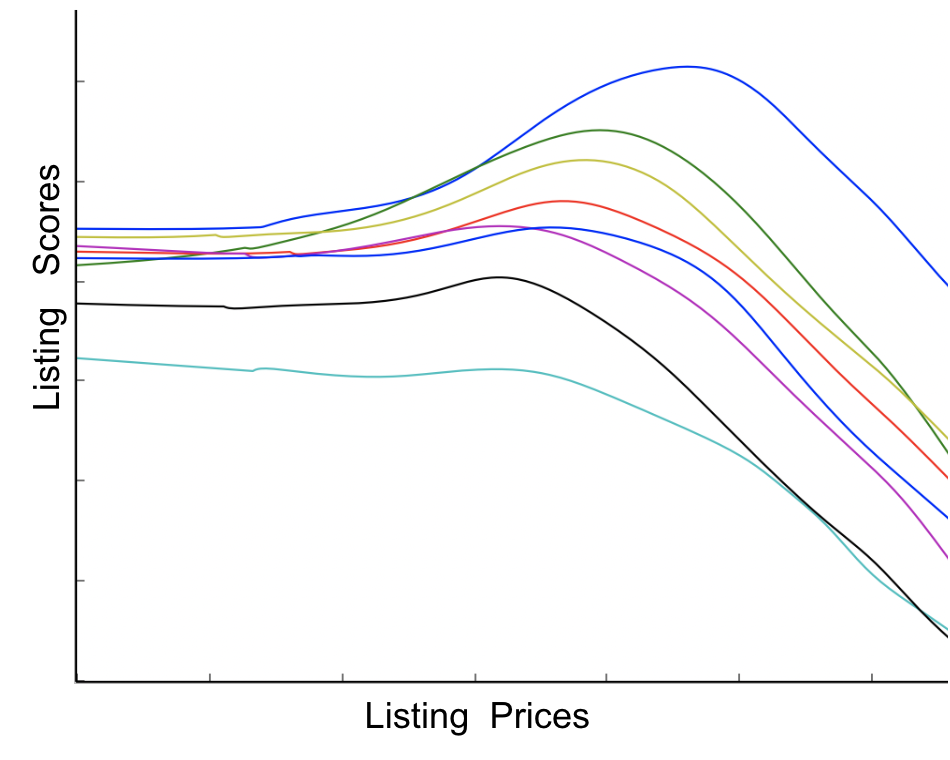}
\caption{ICE curves for price revisited for the two tower architecture.}
\label{fig:iceconehead}
\end{figure}

Revisiting the ICE plots for price, we saw a marked change. Instead of the plots always sloping downwards with price underlining a {\it cheaper is better} interpretation, we saw that the scores peak around certain prices, as shown in Figure~\ref{fig:iceconehead}. This was closer to the {\it right price for the trip} interpretation.

A question raised frequently in this context was whether low quality listings could get up-ranked by the new model simply by targeting a price. Careful inspection of the ICE curve revealed that the score peaks around certain prices were happening only for high quality listings, which were usually ranked near the top to begin with. For most average listings, the plot still maintained a monotonically decreasing curve with respect to price.

\begin{figure}
\includegraphics[height=2.3in, width=2.5in]{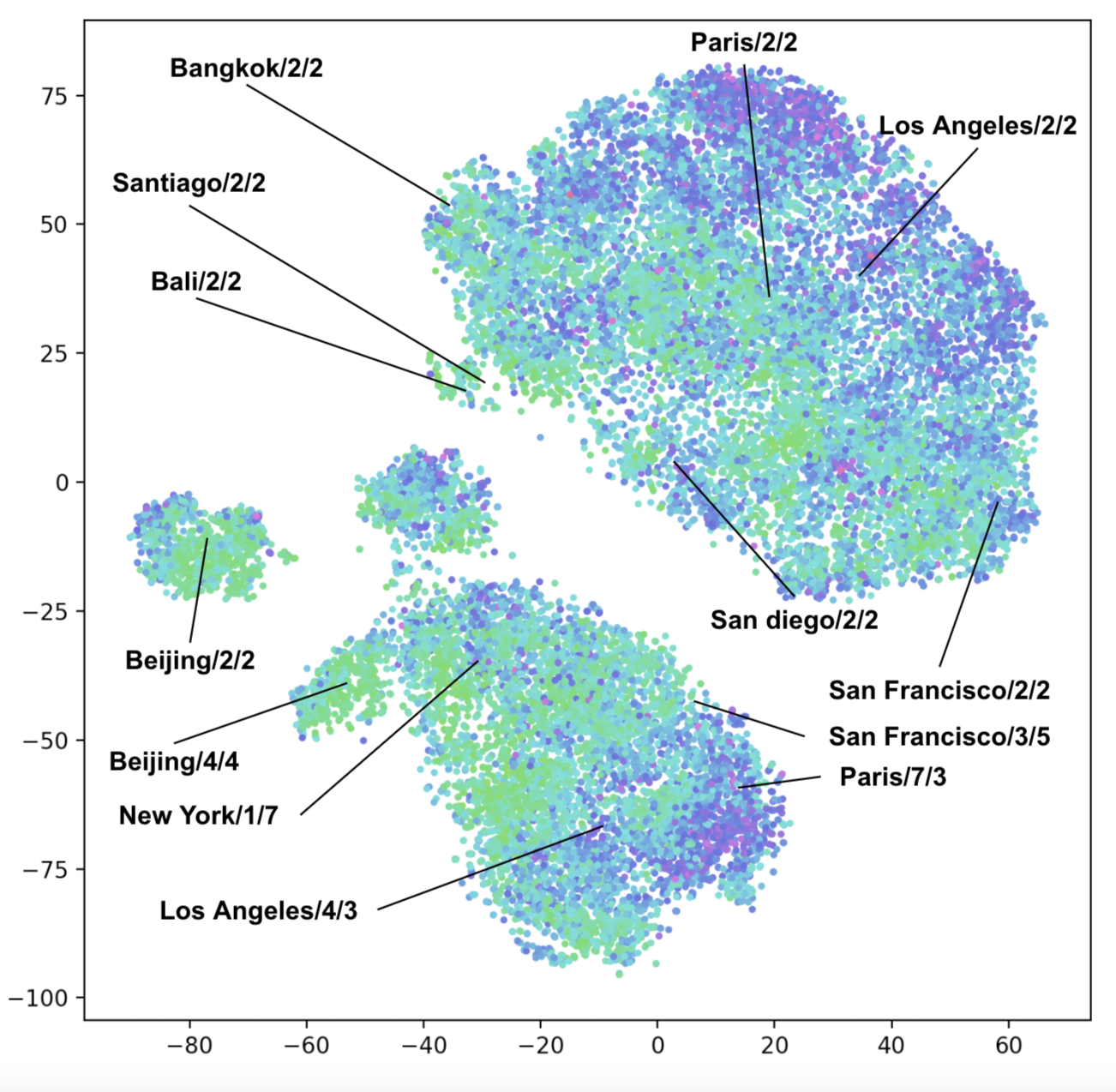}
\caption{t-SNE plot for output of query tower mapped to 2-d. Each dot represents a query. Some of the queries are labeled in the $city/guest\mhyphen count/trip\mhyphen length$ format. Color of the dots indicate price of the listing booked for the query, cheaper towards green, expensive towards blue.}
\label{fig:tsne}
\end{figure}

The notion of the right price and the ideal listing centered around the vector generated by the query tower, so a natural follow up was to investigate exactly what those vectors looked like. For analysis, we ran the two tower DNN on a random sample of searches and collected the output vector of the query tower. Since the $100 \mhyphen d$ vectors were not human interpretable, we applied t-SNE ~\cite{misreadtsne} to reduce them to $2 \mhyphen d$ vectors, which are shown in Figure~\ref{fig:tsne}. Queries corresponding to some of the cities in Figure~\ref{fig:avgcitydiff} are marked on the plot.

It was reassuring to see large clusters forming around similar values of parameters such as guest count and trip length. Within the large clusters, cities that felt intuitively similar were placed comparatively closer to each other.
 
It's worth highlighting that the clusters are not simply price clusters. The price of the booked listing corresponding to the query is represented by the color of the dots, and we see the clusters have colors of all range. While Moscow is typically cheaper than Paris, a booking price in Moscow can easily exceed a booking price in Paris depending on the number of guests, duration of stay, proximity to tourist attractions, weekend vs weekday, and a host of other factors. Price is inextricably linked with all the other dimensions and to grasp the right price for a trip implies a good grasp on all the other factors simultaneously. None of the analysis we did can be used as hard evidence that the two tower architecture had indeed developed this grasp. But the combination of the ICE plots against price, t-SNE visualization of the query tower output, and additional analysis of price movements across cities gave us sufficient confidence that the mechanism was working as intended. 

Laying the series of architectural manipulations to rest, next we move on to addressing a ranking challenge that not only affected guests, but also the other half of the Airbnb community, the hosts.

\section{Improving Cold Start}
In machine learning applications for the travel space, a large fraction of users at any point are new or are using the product after a long gap of time. For all practical purposes users are in a state of continuous cold start as noted in ~\cite{booking}. Handling user level cold start is part of the core ranking formulation itself. So when referring to the cold start problem, we focus our attention on the item level cold start (i.e, how to handle new listings in ranking). As in the case of refining the DNN architecture in Section~\ref{userdriven}, the starting point of our quest was not a literature survey, but the observation of a user problem.

Using NDCG to quantify the position of the booked listings in search results has been the most reliable gauge of model performance for us. Therefore a natural place to investigate user problems was to look for segments of listings where NDCG was lower compared to the overall NDCG. Breaking out the NDCG for booked listings that were new to the platform and comparing them to established listings, we observed a gap of $-6\%$. For context, we have observed statistically significant differences in online bookings from models that differed in NDCG by as little as $0.7\%$. This indicated that the model was making guests work significantly harder to discover the new listings worth booking. To understand this better, we removed all the input features from the DNN that were generated based on historical interactions with guests, such as the number of past bookings for a listing. Removal of these engagement features resulted in a drop of $-4.5\%$ in NDCG. Clearly, the DNN was relying heavily on engagement features. In the absence of these fine grained differentiations from guests for new listings, the DNN was forced to make broad judgements based on the remaining features, fitting close to the average performance of new listings.

\subsection{Approaching Cold Start As Explore-Exploit}
One possible framing of the cold start problem is to consider it a tradeoff between explore and exploit. Ranking strategies could exclusively optimize for bookings in the short term by exploiting knowledge of the current inventory, and betting only on those listings that have a proven track record. But for long term success of the marketplace, it needs to pay some cost to explore the new inventory. This tradeoff can be implemented as an explicit ranking boost for new listings, which allocates higher ranks to new listings than what is determined by the DNN. This allows new listings to collect feedback from guests at a small cost to bookings. The general method has been popular in e-commerce ranking applications, for example in ~\cite{rboost}. The boost can be further refined, capping it by impression counts, or introducing temporal decays. Our first iteration was to test such a boost. Through A/B testing online, we tuned the new listing ranking boost to be bookings neutral compared to no boosting, while allocating $+8.5\%$ additional first page impressions to new listings.

But operating under the explore-exploit paradigm created severe challenges:
\begin{itemize}
\item The new listing ranking boost was pulled in different directions by two opposing forces: 1) a degradation of user experience in the short term due to reduced relevance of the search results (an effect we could measure accurately) vs 2) an improvement in user experience in the long term due to incremental inventory (an effect we found rather difficult to quantify). The lack of a clear and objective definition of the optimal amount of boost lead to vigorous internal debates, with no resolution that satisfied every team interested.
\item Even after arbitrarily fixing an overall budget for the cost of exploration, it became clear that proper use of the budget was dependent on the supply and demand in a particular location. Tolerance for exploration is high when there is high demand, but not as much when demand in a location is scarce. And the need to explore and expand the inventory is high in locations where good supply is constrained. When plenty of high quality listings are lying vacant, there is little incentive to incur the cost of exploration. Supply and demand in turn are governed by location, seasonality and guest capacity among other parameters. So to optimally use the global exploration budget, thousands of localizing parameters were needed, a task impossible to handle manually.   
\end{itemize}

\subsection{Estimating Future User Engagement}
To make the system more manageable, we took a step back and started by asking: what makes a new listing different? The answer, of course, is the absence of user generated engagement features like number of bookings, clicks, reviews etc. Other properties like price, location, amenities are known just like the rest of the listings. In theory, if we had an oracle to predict the engagement features for a new listing with $100\%$ accuracy, it could solve the cold start problem optimally.

So instead of treating cold start as an explore-exploit tradeoff, we reframed it as a problem of estimating the engagement values for a new listing. Reframing the problem unlocked something significant: it allowed us to define an objective ideal for the problem and iteratively work towards it. To solve cold start, we introduced a new component feeding the DNN which predicted the user engagement features for a new listing, both at training and scoring time.

To measure the accuracy of the estimator, we applied the following steps:
\begin{itemize}
\item Sample O(100M) search results from the logs. For each of the search results, randomly sample a listing from the top $100$ positions. These represent a sample of listings that have received ample attention from guests, so have their engagement features sufficiently converged.
\item Let $R_{real}$ denote the rank of the sampled listings obtained from the logs. We denote the rank as real to indicate that the engagement features for the listings are result of real guest interactions. From the rank, we compute real discounted rank as $DR_{real} = log(2.0)/log(2.0 + R_{real})$.
\item Next, for each of the sampled listings, we remove all the engagement features and replace them by the engagement features predicted by the estimator under test. We score the listing with the predicted engagement features, find its new rank in the corresponding logged search result, then compute the discounted rank from it. We denote this by $DR_{predicted}$.
\item For each sampled listing, we compute the error in engagement estimation as $(DR_{real} - DR_{predicted})^2$
\item To get the overall error, we average the error in engagement estimation across all the sampled listings.  
\end{itemize}

The ideal engagement estimator would generate $0$ error. To decide between two estimators, one can pick the estimator with lower error.

For validation we compared two estimation approaches. The baseline was the system used in production which assigned default values for missing features, including engagement features for new listings. The default values were constants crafted by manual analysis of the corresponding features. The comparison was against an estimator which predicted the engagement features by {\it averaging the engagement features of listings geographically nearby to the new listing}. To increase accuracy, it only considered neighboring listings that matched the guest capacity of the new listing, and computed the averages over a sliding time window to account for seasonality. For example, to estimate the number of bookings for a new listing with a two person guest capacity, it took the average number of bookings for all listings within a small radius of the new listing with a capacity of two. This is conceptually similar to the Naive Bayes recommender from ~\cite{coldstart} which used a generative method to estimate the missing information.

\subsection{Test Results}
In offline analysis, the engagement estimator described above reduced engagement estimation error by $-42\%$ when compared to using default values. 

In online A/B experiment, we observed an improvement of $+14\%$ in bookings of newly created listings, along with a $+14\%$ increase in share of impressions of first page results. Apart from its impact on new listings, overall bookings increased by $+0.38\%$, indicating an overall improvement in the user experience. 

From examining challenges with the data feeding the DNN, we transition to problems surrounding how the DNN interpreted the data presented, and the issue of positional bias.

\section{Eliminating Positional Bias}
The starting point for our investigation into positional bias was something quite unrelated. Similar to the observation around lower NDCG for new listings, another segment that showed lower than expected performance was boutique hotels and traditional bed and breakfasts, a segment that was growing rapidly as part of the inventory. One hypothesis coming out of the observation was that inventory historically under-represented in the training data were not ranked optimally due to positional bias. But unlike the link between new listings performance and cold start, there was no strong reason to believe positional bias was the sole culprit in this case; there were multiple other hypothesis. While we found focusing on user problems a much better approach than simply importing ideas from literature surveys, this by itself was not a panacea. Establishing a causal link between a user problem and a shortcoming in the model was far from straightforward. In the current scenario, we were shooting in the dark. But while at it, we decided to go after the biggest gaps in modeling that explained the observations. And a literature survey was crucial in identifying where major gaps might be lurking in our model.

\subsection{Related Work}
Given a user $u$ who issues a query $q$, the probability of the user booking a listing $l$ from the search results can be decomposed into two factors:
\begin{itemize}
\item The probability the listing was relevant to the user. This probability can be represented as $P(relevant=1|l, u, q)$ to make explicit the dependencies on the listing, user and query.
\item The probability that the user examined the listing given it was at position $k$ in the search result. This may depend on the user (e.g. users on mobile may have higher bias for the top results) or on the query (e.g. users with short lead days may pay even less attention to the bottom results). We represent this probability as $P(examined=1|k, u, q)$, independent of the listing $l$. The influence of the listing on the booking event is completely accounted for by $P(relevant=1|l, u, q)$
\end{itemize}

Using the simplifying assumptions of the position based model described in ~\cite{positionbasedmodel}, we represent the probability of the user booking a listing simply as a product of the two decomposed probabilities.
\begin{equation}
P_{booking} = P(relevant=1|l, u, q) * P(examined=1|k, u, q)
\end{equation}

By directly training a model to predict bookings, the model learns to predict $P_{booking}$ which is dependent on $P(examined=1|k, u, q)$. That in turn depends on the position $k$, a decision taken by the previous ranking model. The current model becomes dependent on previous models as a result.

Ideally we would like the model to focus exclusively on $P(relevant=1|l, u, q)$ and rank listings by relevance alone. To achieve that, ~\cite{Joachims:2017} describes a method with two key concepts:
\begin{itemize}
\item A {\it propensity model} to predict $P(examined=1|k, u, q)$.
\item Weighing each training example by the inverse of the predicted propensity.
\end{itemize} 
While constructing the propensity model typically involves perturbing the search results to collect examples of counterfactuals, ~\cite{Agarwal} describes methods to construct the propensity model without additional interventions.

\subsection{Position As Control Variable}
Our solution has two key highlights. Firstly, it is non-intrusive and does not need any randomization of the search results. We rely on some unique properties of search results at Airbnb which make listings appear at different positions, even when their corresponding scores while ranking are more or less invariant:
\begin{itemize}
\item Listings represent physical entities that can be booked only once for a given date range. As listings get booked and disappear from search, it shifts the positions of the remaining listings.
\item Each listing has its own unique calendar availability, so different listings get to appear at different positions for similar queries across date ranges.
\end{itemize}

The second highlight of our solution is that we do not build an explicit propensity model. Instead, {\it we introduce position as a feature in the DNN, regularized by dropout}. During scoring we set the position feature to 0. The rest of the section describes the intuition behind why this works.

We take the DNN described in Section ~\ref{twotower} as the foundation, with query, user and listing features as inputs. Using the notation $q$ (query features), $u$ (user features), $l$ (listing features), $\theta$ (DNN parameters), we express the output of the DNN as 
\begin{equation}
dnn_{\theta}(q, u, l) = rel_{\theta}(q, u, l) * pbias_{\theta}(q, u, l)
\end{equation}
mirroring the assumption made by the position based model in ~\cite{positionbasedmodel}.  
Here $rel_{\theta}(q, u, l)$ estimates $P(relevance=1|l, u, q)$ which we refer to as the relevance prediction. And $pbias_{\theta}(q, u, l)$ estimates $P(examination=1|k, u, q)$ which we call the positional bias prediction.

It becomes immediately apparent that $pbias_{\theta}(q, u, l)$ is missing the position of the listing $k$ as input since the quantity it is trying to estimate is dependent on $k$. So our first step is to add $k$ as an input feature to the DNN. Since both the relevance prediction and the positional bias prediction are fed by the DNN inputs, adding $k$ to the inputs transforms our representation of the DNN to 
\begin{equation}
dnn_{\theta}(q, u, l, k) = rel_{\theta}(q, u, l, k) * pbias_{\theta}(q, u, l, k)
\end{equation}

Given that $P(examined=1|k, u, q)$ is independent of $l$, any dependence of the positional bias prediction on $l$ can be treated as an error. We assume that with sufficient amount of training data, the learnt parameters $\theta$ are able to minimize that error and the positional bias prediction becomes independent of $l$ for all practical purposes. We capture this assumption as 
\begin{equation}
dnn_{\theta}(q, u, l, k) = rel_{\theta}(q, u, l, k) * pbias_{\theta}(q, u, k) \label{dnneq}
\end{equation}
dropping $l$ from $pbias_{\theta}(q, u, l, k)$.

While scoring, we set the position feature $k$ to $0$. Within a given search, $q$ and $u$ are invariant across the listings scored by the DNN. We use $Q$ and $U$ to represent the query and user features for a particular search. The position bias prediction therefore becomes $pbias_{\theta}(Q, U, 0)$, which is an invariant for all the listings in a particular search result. Naming the invariant $\beta$, equation ~\eqref{dnneq} at scoring time can be rewritten as
\begin{equation}
dnn_{\theta}(Q, U, l, 0) = rel_{\theta}(Q, U, l, 0) * \beta
\end{equation}
This makes the comparison of two listing scores independent of positional bias and dependent only on listing relevance. In essence, we added position as a control variable ~\cite{controlvar} in the ranking model.

\subsection{Position Dropout}
Under the position based model assumption, adding position as a control variable effectively eliminates the position bias prediction from listing rankings, but it introduces a new problem. The relevance prediction is now dependent on position as a feature. This runs the risk of the DNN relying on the position feature during training to predict relevance, but not able to utilize that learning while scoring where the position feature is always set to $0$. Comparing the NDCG of the DNN with position as feature to the baseline without the position feature, we see a drop of around $-1.3\%$. So a straightforward introduction of the position as a control variable seem to hurt the relevance prediction.

To reduce the dependence of the relevance prediction on the position feature, we regularize it down using dropout ~\cite{dropout}. During training, we probabilistically set the position for a listing to $0$, controlled by the dropout rate.

\begin{figure}
\includegraphics[height=1.7in, width=2.3in]{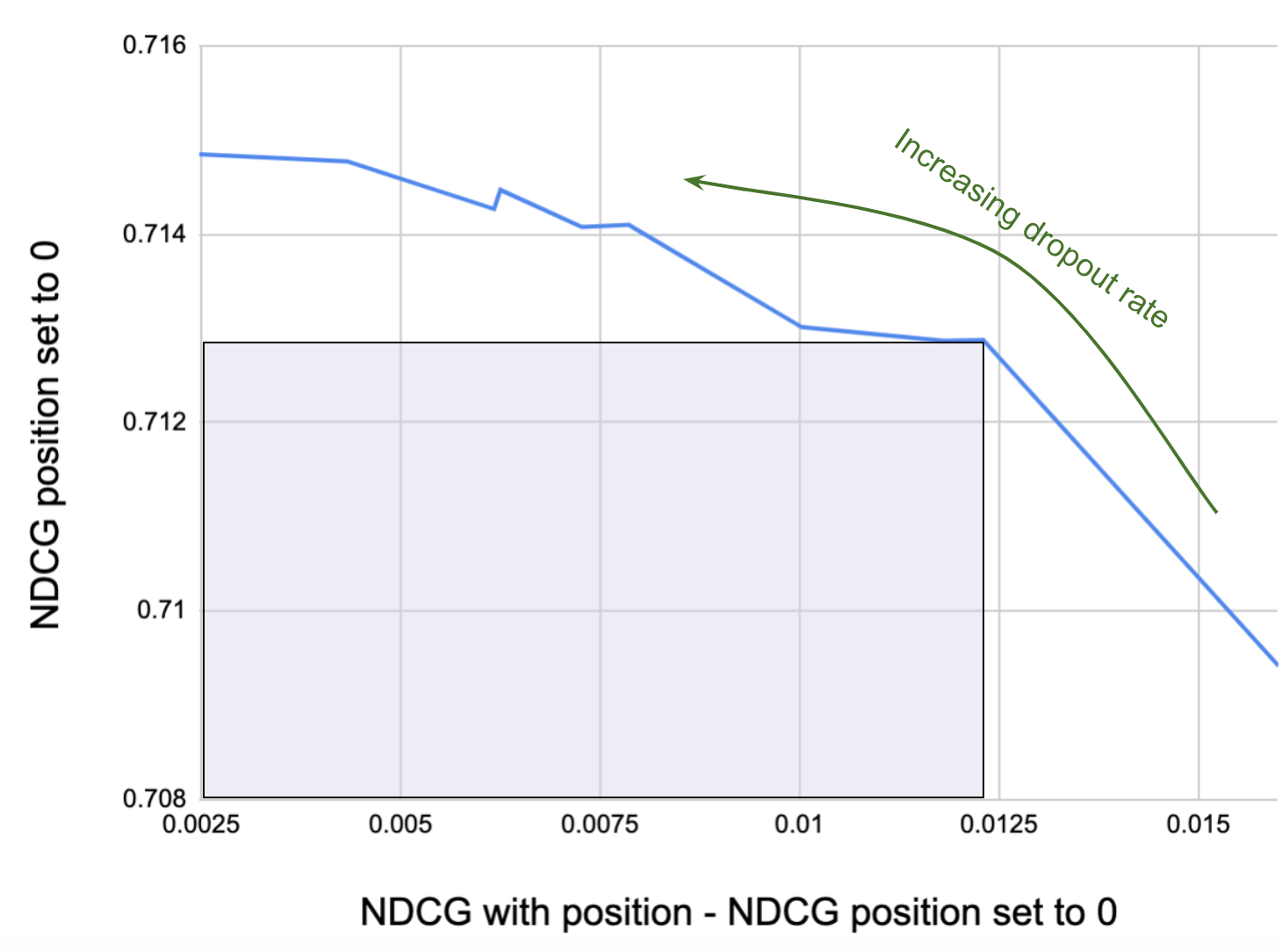}
\caption{Y-axis is NDCG with position set to $0$. X-axis is difference between the NDCG with position set to $0$ and position kept as is.}
\label{fig:drate}
\end{figure}

The dropout rate presents a tradeoff between noise-free access to the position feature to infer positional bias accurately vs making the position feature noisy to regularize it away from relevance prediction. We try to find a balance for the tradeoff through the following steps:
\begin{itemize}
\item Sweep through the range of dropout rates and compute two flavors of NDCG on a test set. First one by setting the position to $0$ during test. This measures the relevance prediction and denoted $NDCG_{rel}$. The second one by keeping the position feature which measures the combined relevance and position bias prediction, denoted $NDCG_{rel+pbias}$.
\item Subtract $NDCG_{rel+pbias}-NDCG_{rel}$ to get a measure of the positional bias prediction. The intuition here is that by comparing the quality of ranking with and without the position input, we get an estimate of the contribution of position towards ranking. Plot it against $NDCG_{rel}$ to obtain the curve in Figure~\ref{fig:drate}.
\item To balance between the relevance prediction and positional bias prediction, pick a point on the curve where the positional bias prediction is sufficiently advanced on the x-axis, without incurring too much of a drop in relevance prediction on the y-axis.
\end{itemize}
Through this exercise we ended up selecting a dropout rate of $0.15$.

\subsection{Test Results}
We tested the idea by an online A/B test, where the control was the DNN from Section ~\ref{twotower} with no notion of positional bias. The treatment was the same DNN, but trained with position as a feature regularized by a dropout rate of $0.15$. In the online test we observed a gain of $+0.7\%$ in bookings. 

Alongside the bookings gain, a lift of $+1.8\%$ in revenue was a pleasant surprise. The revenue side effect illustrated how positional bias had built up over multiple iterations of the model. For the ranking model, it's relatively easy to learn the effect of price as it comes as a very clean feature and the data strongly suggests a preference for lower prices. The balancing forces of quality, location, etc. are much harder to learn. As a result, initial simplistic models heavily relied on lower prices. Over multiple model iterations we improved our understanding of quality and location, but by then the bias towards cheaper prices was already ingrained in the training data. This stickiness made successive models overestimate the preference for lower prices. Eliminating positional bias allowed the model to get closer to the true preference of guests, and strike a more optimal balance between price, quality and location. The revenue lift observed was a direct fallout of that. Finally, to close the loop on where we started, we observed a $+1.1\%$ increase in bookings for boutique hotels.

\section{Conclusion}
Deep learning continues to flourish in search ranking at Airbnb. We feel genuine gratitude towards the community for providing the deep learning ecosystem, for the open exchange of ideas, and for the opportunity to join the conversation by sharing our own experiences. But the highlight of our journey is the realization that to push the boundaries of our DNNs, the inspiration was not going to come from some external source. For that we had to follow the lead of our users.


\bibliographystyle{ACM-Reference-Format}

\balance 
\bibliography{kdd-bibliography}

\end{document}